\documentclass{article}
\usepackage{ijcai26}
\usepackage{times}
\usepackage{soul}
\usepackage{url}
\usepackage[hidelinks]{hyperref}
\usepackage[utf8]{inputenc}
\usepackage[small]{caption}
\usepackage{graphicx}
\usepackage{amsmath}
\usepackage{amssymb}
\usepackage{amsthm}
\usepackage{booktabs}
\usepackage{algorithm}
\usepackage{algorithmic}
\usepackage{multirow}
\usepackage[switch]{lineno}
\usepackage{pifont}
\usepackage{threeparttable}
\usepackage{makecell}
\usepackage{array,multirow,graphicx}
\usepackage{mydef}
\usepackage{tabularx}
\usepackage[T1]{fontenc}
\newtheorem{definition}{Definition}

\title{Dynamic Heterogeneous Graph Representation Learning: A Survey}

\author{
Huan Liu$^1$
\and
Pengfei Jiao$^1$\footnote{Corresponding author.} \and
Jie Yin$^{2}$ \and
Hongjiang Chen$^1$ \And
Zhidong Zhao$^{1,3}$ 
\affiliations
$^1$School of Cyberspace, Hangzhou Dianzi University, China\\
$^2$Discipline of Business Analytics, The University of Sydney, Australia\\
$^3$Zhejiang Provincial Key Laboratory for Sensitive Data Security Protection and Confidentiality Management, Hangzhou, China
\emails
\{huanliu, pjiao, hchen, zhaozd\}@hdu.edu.cn,
jie.yin@sydney.edu.au
}

\begin{document}

\maketitle

\begin{abstract}
Graph representation learning (GRL) serves as a canonical paradigm for modeling complex networks. However, real-world AI systems inherently manifest as evolving heterogeneous entities with complex interactions, posing significant challenges to static or homogeneous modeling. To address these complexities, representation learning for Dynamic Heterogeneous Graphs (DHGs) has emerged as a vital approach for learning low-dimensional representations that simultaneously preserve structural semantics and temporal dynamics. This survey presents the first systematic review of DHG representation learning methods. We first introduce a unified formal definition that encompasses both discrete-time and continuous-time DHGs from the perspective of temporal granularity. Building upon this formulation, we propose a novel algorithm-centric taxonomy that categorizes existing literature, including early embedding-based approaches, graph neural network (GNN)-based models, and relatively recent Transformer-based DHG methods, while explicitly highlighting their intrinsic modeling biases with respect to dynamic granularity. Furthermore, we summarize representative applications of DHG representation learning, along with commonly used datasets and benchmarks. Finally, we discuss promising research directions that guide future advances in this rapidly evolving field.
\end{abstract}

\section{Introduction}

Graphs have been established as a foundational data structure for modeling complex real-world networks and have seen widespread applications across many fields in recent years, such as social network analysis~\cite{li_thgnn_2023,wang_htgformer_2025}, foundation models~\cite{fey_relational_2024,liu_graph_2025_tpami}, and recommendation systems~\cite{liu_mg-dvd_2021,li_periodic_2024}. By preserving the complex structure and dependencies of graphs in a low-dimensional vector space, graph representation learning (GRL) has been a cornerstone for enabling generalizable reasoning over relational data in modern intelligent systems such as large language models (LLMs), and has attracted considerable research interest~\cite{fey_relational_2024,jin_llm_graphs_2024,liu_graph_2025_tpami}.

However, real-world systems typically manifest as dynamically evolving, multi-typed entities with complex interactions~\cite{hong_heteta_2020,qi_athitd_2025}, where dynamics and heterogeneity are inseparably coupled rather than independent properties~\cite{xu_heterogeneous_2022,wang_tesa_2025}. Thus, Dynamic Heterogeneous Graph (DHG) representation learning has emerged as a vital paradigm for capturing the joint interplay between heterogeneous semantics and temporal evolution~\cite{fey_relational_2024,liu_graph_2025_tpami}. Nevertheless, existing research typically models heterogeneity and dynamics separately~\cite{wang2022survey,kazemi2020representation}, and effective DHG representation learning remains non-trivial due to several key challenges. First, heterogeneous semantics and temporal dynamics are intrinsically coupled; thus, methods that simply concatenate heterogeneous modeling with sequential models fail to capture relation-specific evolutionary behaviors. Second, as heterogeneous interactions follow different temporal behaviors, homogeneous modeling is insufficient to capture the sudden, periodic, and long-term cumulative dynamic changes of multi-typed nodes/edges. Third, discrete- and continuous-time DHGs exhibit distinct structural evolution patterns at different temporal granularities; while the former captures coarse-grained global updates, the latter tracks rapid, asynchronous, and localized changes. Consequently, designing unified architectures that preserve granularity-specific dynamics while effectively filtering temporal noise remains an open challenge. Finally, explicitly parameterizing relation-specific semantics and dynamics increases computational overhead, thus hindering the scalability of large-scale DHG learning.

Although a growing body of studies has addressed certain facets of these challenges~\cite{dileo_durendal_2023,wang_htgformer_2025}, the literature remains fragmented and lacks a unified review. To bridge this gap, we present the first comprehensive survey dedicated to DHG representation learning. We establish a unified formal definition of DHGs that bridges discrete-time and continuous-time formulations. On this basis, we propose an algorithm-centric taxonomy of DHG representation learning, categorizing methods into embedding-based, GNN-based, and Transformer-based paradigms, and explicitly characterizing their inductive biases under different temporal granularities. We further discuss representative applications and summarize commonly used open-source datasets and benchmarks for DHG representation learning, revealing widespread inconsistencies in graph construction and fixed temporal partitions that hinder fair comparison and reliable performance evaluation.

\begin{figure*}[ht]
\centering
\begin{forest} dhg-e
  [DHG Representation Learning,
    rootnode, rotate=90, anchor=north, fill=mygreen, draw=black
    [Embedding-based, levelone, fill=myred
      [Random Walk-based, leveltwo, fill=myred
        [Change2vec~\cite{bian_network_2019}{,} DyHINE~\cite{xie_learning_2021}{,} H2TNE~\cite{bai_h2tne_2023},
         methods, fill=myred]
      ]
      [Incremental Update-based, leveltwo, fill=myred
        [DyHNE~\cite{wang_dynamic_2022}{,} M-DHIN~\cite{fang_scalable_2022}{,} LIME~\cite{peng_lime_2022},
         methods, fill=myred]
      ]
      [Temporal Point\\Process-based, leveltwo, fill=myred
        [HDGAN~\cite{li_heterogeneous_2020}{,} HPGE~\cite{ji_dynamic_2021}{,}  \\
         THINE~\cite{huang_temporal_2021}{,} SemE~\cite{zhou_temporal_2023}{,} TeSa~\cite{wang_tesa_2025},
         methods, fill=myred]
      ]
    ]
    [GNN-based, levelone, fill=myyellow
      [Relation-specific, leveltwo, fill=myyellow
        [DyHAN~\cite{yang_dynamic_2020}{,} HTGNN~\cite{fan_heterogeneous_2022}{,} HTHGN~\cite{liu_heterogeneous_2025},
         methods, fill=myyellow]
      ]
      [Meta-structure-guided, leveltwo, fill=myyellow, 
        [MG-DVD~\cite{liu_mg-dvd_2021}{,} DHANE~\cite{li_dynamic_2024},
         methods, fill=myyellow]
      ]
      [Recurrent-based, leveltwo, fill=myyellow
        [DyHATR~\cite{hutter_modeling_2021}{,} DURENDAL~\cite{dileo_durendal_2023}{,} HGN2T~\cite{liu_hgn2t_2024},
         methods, fill=myyellow]
      ]
    ]
    [Transformer-based, levelone, fill=mypurple
      [Structure-oriented, leveltwo, fill=mypurple
        [HGT~\cite{hu_heterogeneous_2020}{,} HT-Trans~\cite{9892546}{,} \\
        THGAT~\cite{zhang_dynamic_2023}{,} DHGAS~\cite{zhang_dynamic_2023},
         methods, fill=mypurple]
      ]
      [Interaction-oriented, leveltwo, fill=mypurple
        [STHN~\cite{li_simplifying_2023}{,} MIGNN~\cite{yue_unified_2025},
         methods, fill=mypurple]
      ]
      [LLM-enhanced, leveltwo, fill=mypurple
        [CasMLN~\cite{wang_llm-enhanced_2024}{,} HTGformer~\cite{wang_htgformer_2025},
         methods, fill=mypurple]
      ]
    ]
    [Application-oriented, levelone, fill=myblue
      [Cybersecurity, leveltwo, fill=myblue
        [TimeSAGE~\cite{shekhar_entity_2020}{,} HTGT~\cite{fan_heterogeneous_2021}{,} ATHITD~\cite{qi_athitd_2025},
         methods, fill=myblue]
      ]
      [Traffic Forecasting, leveltwo, fill=myblue
        [HetETA~\cite{hong_heteta_2020}{,} REGNN~\cite{luo_dynamic_2020}{,} STHGFormer~\cite{li_towards_2024},
         methods, fill=myblue]
      ]
      [Recommendation, leveltwo, fill=myblue
        [DHIMN~\cite{xie_sequential_2021}{,} SUPA~\cite{wu_instant_2023}{,} DHGP~\cite{li_periodic_2024},
         methods, fill=myblue]
      ]
      [Multimodal, leveltwo, fill=myblue
        [HSSHG~\cite{wang_hsshg_2024}{,} DHGRNN~\cite{wang_two-stream_2025}{,} HDGR~\cite{dai_cross-modal_2025},
         methods, fill=myblue]
      ]
      [Information Diffusion, leveltwo, fill=myblue
        [
        HINTS~\cite{jiang_hints_2021}{,} SI-HDGNN~\cite{xu_heterogeneous_2022}{,} THGNN~\cite{li_thgnn_2023},
         methods, fill=myblue]
      ]
    ]
  ]
\end{forest}
\caption{A Taxonomy of Dynamic Heterogeneous Graph (DHG) Representation Learning Methods.}
\label{fig:taxonomy}
\vspace{-3mm}
\end{figure*}
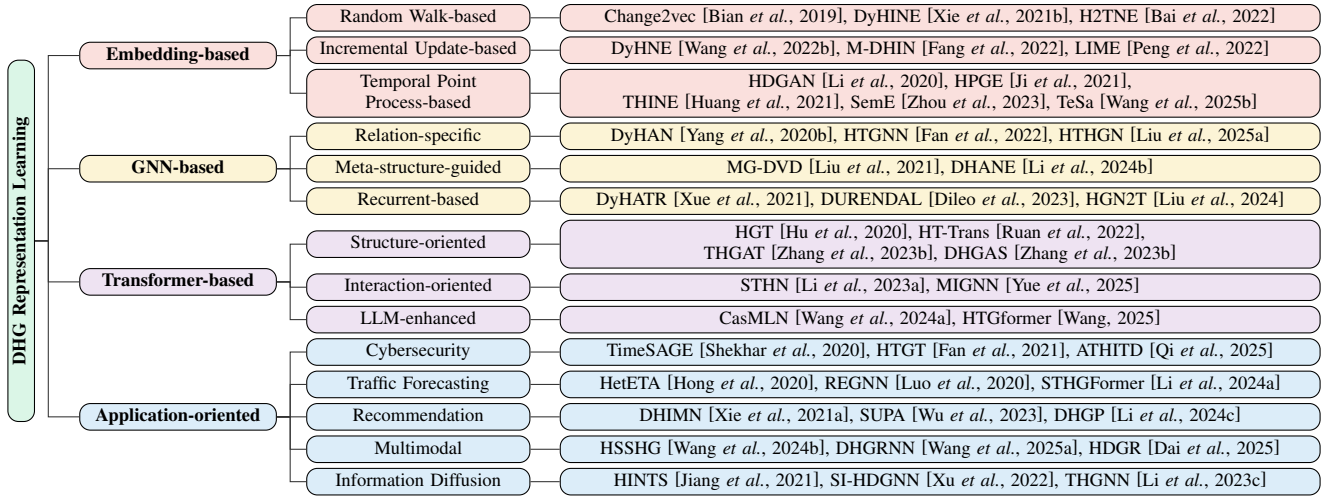

Distinct from prior GRL surveys, this work provides a unified and systematic review of DHG representation learning. Existing surveys on heterogeneous graphs focus on heterogeneous modeling principles and representation techniques under static settings~\cite{yang2020heterogeneous,wang2022survey}. In parallel, surveys on dynamic GRL offer valuable insights into temporal modeling paradigms and evaluation protocols~\cite{kazemi2020representation,10.1145/3483595}, but primarily under homogeneous assumptions. This work also diverges from benchmarking studies such as TGB~2.0~\cite{gastinger_tgb_2024}, which emphasize evaluation frameworks rather than algorithmic mechanisms. While these existing efforts have significantly advanced their respective areas, the joint modeling of heterogeneity and temporal evolution remains under-explored, necessitating a dedicated and holistic review of DHG representation learning. Our major contributions include:
\begin{itemize}
    \item We present a formal definition of DHG encompassing both discrete-time and continuous-time formulations, and introduce a structured taxonomy organizing methods into four families based on their core architecture. An overview is presented in Figure~\ref{fig:taxonomy}.
    \item We systematically review representative methods within each family, analyzing their technical mechanisms and identifying shared limitations that constrain current DHG learning. We further compile open-source datasets and benchmarks, exposing inconsistent construction and deterministic evaluation protocols that hinder fair comparison and reliable assessment.
    \item We identify three progressive research frontiers toward DHG foundation models: efficiency for enabling scaling law investigation, generalizability for cross-domain transfer and continual learning, and trustworthiness for enabling explainable DHG learning and causal discovery.
\end{itemize}


\tikzset{
  nodebase/.style={
    draw=black,
    rounded corners,
    minimum height=1em,
    text opacity=1,
    align=center,
    text=black,
    font=\scriptsize,
    inner xsep=3pt,
    inner ysep=1pt,
    fill opacity=.85,
  },
  rootnode/.style={
    nodebase,
    text width=13em,     
    font=\scriptsize\bfseries,
  },
  levelone/.style={
    nodebase,
    text width=6.8em,     
    font=\scriptsize\bfseries,
  },
  leveltwo/.style={
    nodebase,
    text width=7.8em,    
  },
  methods/.style={
    nodebase,
    text width=28em,   
  }
}

\forestset{
  dhg-e/.style={
    for tree={
        forked edges,
        grow=east,
        reversed=true,
        anchor=base west,
        parent anchor=east,
        child anchor=west,
        base=middle,
        font=\scriptsize,
        rectangle,
        draw=black,
        edge=black!80,
        rounded corners,
        align=center,
        minimum width=2em,
        s sep=1.8pt,
        inner xsep=3pt,
        inner ysep=1pt
    },
    where level=3{font=\scriptsize}{},
    where level=4{font=\scriptsize}{},
    where level=5{font=\scriptsize}{},
  }
}

\section{Preliminaries and Notations}
This section introduces the formal definition for DHGs and the representation learning tasks.

\begin{definition}[Dynamic Heterogeneous Graph]
A \emph{Dynamic Heterogeneous Graph (DHG)} is defined as a graph jointly modeling structural heterogeneity and temporal evolution:
\[
\mathcal{G} = (\mathcal{V}, \mathcal{E}, \mathcal{A}, \mathcal{R}, \mathcal{T}, \phi, \varphi),
\]
where $\mathcal{V}$ and $\mathcal{E}$ denote the (possibly time-varying) sets of nodes and edges over a temporal domain $\mathcal{T}$. 
The type-mapping functions $\phi:\mathcal{V}\rightarrow\mathcal{A}$ and $\varphi:\mathcal{E}\rightarrow\mathcal{R}$ assign each node and edge to a node-type set $\mathcal{A}$ and an edge-type set $\mathcal{R}$, respectively.
\end{definition}

\textbf{Remark.} This formulation is general and subsumes multiple mainstream graph types as special cases:

\emph{Heterogeneous vs.\ Homogeneous.}
A graph $\mathcal{G}$ is \emph{heterogeneous} if it involves multiple semantic types, i.e.,
$|\mathcal{A}| + |\mathcal{R}| > 2$.
When $|\mathcal{A}| = 1$ and $|\mathcal{R}| = 1$, $\mathcal{G}$ degenerates to a \emph{homogeneous graph} with a single node and edge type.

\emph{Discrete vs.\ Continuous.}
The temporal domain $\mathcal{T}$ governs the dynamic nature of the graph.
If $|\mathcal{T}| = 1$, $\mathcal{G}$ reduces to a \emph{static graph} $\mathcal{G}=(\mathcal{V},\mathcal{E},\mathcal{A}, \mathcal{R}, \phi, \varphi)$ representing a single heterogeneous graph snapshot.
If $\mathcal{T}$ is a discrete ordered set, then $\mathcal{G}$ is a \textit{discrete-time} DHG, represented as a sequence of heterogeneous graph snapshots $\mathcal{G} = \{\mathcal{G}^{(t)}\}_{t\in\mathcal{T}}$.
If $\mathcal{T}$ is continuous, the graph is a \textit{continuous-time} DHG consisting of a sequence of timestamped edges:
\[
\mathcal{E}=\{(u,v,r,t)\mid u,v\in\mathcal{V},\, r\in\mathcal{R},\, t\in\mathcal{T}\},
\]
thereby capturing fine-grained, asynchronous interactions.

\begin{definition}[DHG Representation Learning]
Given a DHG $\mathcal{G} = (\mathcal{V}, \mathcal{E}, \mathcal{A}, \mathcal{R}, \mathcal{T}, \phi, \varphi)$ with optional node features $\mathbf{X}^{(t)}$, the goal of \emph{DHG representation learning} is to learn an encoder $f_\Theta: \mathcal{G} \rightarrow \mathbb{R}^{|\mathcal{V}| \times d}$ that maps each node $v \in \mathcal{V}$ to a low-dimensional embedding $\mathbf{h}_v^{(t)} \in \mathbb{R}^d$, such that $\mathbf{h}_v^{(t)}$ jointly preserves heterogeneity and dynamics proximity, supporting various downstream tasks.
\end{definition}

\section{DHG Representation Learning Taxonomy}

DHG representation learning requires jointly encoding type-specific semantics and temporal dependencies, as different relation types exhibit distinct evolutionary patterns. Based on the algorithmic mechanism for capturing such coupled dynamics, we categorize methods into: (1) embedding-based approaches preserving structural and temporal proximity through random walks or matrix factorization; (2) GNN-based approaches with type-aware message-passing and aggregation across snapshots; (3) Transformer-based approaches unifying type and temporal signals within attention mechanisms; and (4) application-oriented approaches formulating domain problems as DHG learning tasks. Table~\ref{tab:method_comparison} summarizes representative methods.

\subsection{Embedding-based}
Embedding-based approaches learn node representations by preserving structural proximity derived from neighborhoods. Based on temporal granularity, we distinguish: (1) random walk-based and (2) incremental update-based methods, both operating on discrete snapshots to capture co-occurrence patterns; and (3) temporal point process-based methods modeling continuous-time events and fine-grained interactions.

\subsubsection{Random Walk-based}
Random-walk-based methods leverage predefined meta-structure to capture the co-evolution of network topology and heterogeneous semantics across temporal snapshots.
The core objective is to maximize the co-occurrence likelihood of nodes in sequences generated by constrained random walks:
\begin{equation}
\mathcal{L}
= \sum_{v \in V^{(t)}} \sum_{c \in \mathcal{N}_\mathcal{S}^{(t)}(v)}
\log P(c \mid v;\theta),
\end{equation}
where $\mathcal{S} \in \{\mathcal{P},\mathcal{M}\}$ denotes the meta-paths or meta-graphs and $\mathcal{N}_{\mathcal{S}}^{(t)}(v)$ represents the time-aware context of node $v$ induced by meta-structure $\mathcal{S}$ at snapshot $t$. Foundational approaches like Change2vec~\cite{bian_network_2019} perform walks on historical-current snapshot pairs to preserve short-term temporal continuity. To capture complex dependencies, DyHINE~\cite{xie_learning_2021} and M-DHIN~\cite{fang_scalable_2022} introduce attention mechanisms and meta-graph guidance to weight relevant paths dynamically. Recent advances, such as H2TNE~\cite{bai_h2tne_2023}, employ hyperbolic embeddings to capture hierarchical scale-free DHG structures.

While effective in preserving structural proximity, these methods rely on expert-defined meta-paths/graphs and often struggle to incorporate rich node attributes, limiting their expressiveness for complex semantic tasks.

\subsubsection{Incremental Update-based}
Incremental update frameworks efficiently model network evolution by focusing on localized calibration rather than retraining. The core of these methods is to combine the previous state with DHG perturbations to update node representations:
\begin{equation}
\mathbf{h}_u^{(t)} = f_{\text{upd}}\!\left(\mathbf{h}_u^{(t-1)}, \Delta \mathcal{G}^{(t)}\right).
\end{equation}
DyHNE~\cite{wang_dynamic_2022} models evolution as perturbations of meta-path augmented adjacency matrices, efficiently updating embeddings via generalized eigenvalue perturbation.
LIME~\cite{peng_lime_2022} employs a recursive neural network to selectively adjust shared semantic embeddings within a local cuboid space.
Besides, Change2vec~\cite{bian_network_2019}, DyHINE~\cite{xie_learning_2021} and M-DHIN~\cite{fang_scalable_2022} explicitly identify nodes affected by edge additions/deletions and node insertions/removals and restrict embedding updates to local neighborhoods rather than the entire network.

Despite their scalability advantages, these methods suffer from cumulative approximation drift over long temporal horizons and struggle with abrupt structural changes that violate temporal smoothness assumptions.

\subsubsection{Temporal Point Process-based}
Temporal point process (TPP) frameworks treat DHG evolution as a stream of asynchronous events, modeling the conditional intensity of heterogeneous interactions.
The technical core involves modeling the conditional intensity function $\lambda(e)$ of a heterogeneous event $e = (u, v, r, t)$, which characterizes the instantaneous arrival rate of a type-$r$ relation between source node $u$ and target node $v$ at time $t$:
\begin{equation}
    \lambda(e) = \mu_r(u, v) + \sum_{e_h: t_h < t} \alpha(e_h, e) \cdot \kappa(t - t_h),
\end{equation}
where $\mu_r(\cdot)$ represents the base rate, $\alpha(\cdot)$ measures the excitation influence from a historical event $e_h$, and $\kappa(\cdot)$ is a kernel function capturing the temporal decay influence. HPGE~\cite{ji_dynamic_2021} integrates a heterogeneous evolving attention mechanism to distinguish fine-grained excitation patterns between different-typed historical and current events, while utilizing temporal importance sampling to extract representative interactions efficiently. HDGAN~\cite{li_heterogeneous_2020} and THINE~\cite{huang_temporal_2021} further leverage meta-paths to constrain candidate sets and employ hierarchical attention to jointly model semantic relevance and structural influence. SemE~\cite{zhou_temporal_2023} abstracts chronological meta-path instances as semantic units driven by an attention-Hawkes process. Besides, T\textsc{e}S\textsc{a}~\cite{wang_tesa_2025} employs a trajectory-based neural TPP to encode node interaction sequences independently, enhancing scalability and long-term pattern capture.

Despite computational efficiency, embedding-based methods remain constrained by predefined proximity measures that limit representation capacity, shallow architectures that preclude higher-order structural modeling, and the absence of task-specific gradients that impedes downstream transfer.

\subsection{GNN-based}
GNN-based methods extend message-passing to DHGs by incorporating heterogeneous edge types and temporal evolution, predominantly operating at the snapshot granularity. Based on the mechanism for integrating type-aware propagation, we categorize these methods into: (1) relation-specific GNNs that decompose each snapshot into type-specific subgraphs with separate transformation matrices; (2) meta-structure-guided GNNs that constrain propagation along predefined meta-paths or meta-graphs to capture high-order semantic patterns; and (3) recurrent-based GNNs that couple spatial aggregation with temporal state evolution through RNN units.

\subsubsection{Relation-specific GNNs}
Relation-specific GNNs represent a dominant paradigm for modeling DHGs by adopting a divide-and-conquer strategy, decomposing complex heterogeneous interactions into multiple relation-specific subgraphs across different snapshots. These methods typically employ hierarchical message-passing and aggregation to disentangle the influence of different relation types while maintaining temporal consistency:
\begin{align}
\mathbf{h}_v^{(t)} = \text{AGG}_{\text{time}} \left( \left\{ \text{AGG}_{\text{type}} \left( \{ \mathbf{h}_u^{(t)} \}_{u \in \mathcal{N}_r^{(t)}(v)} \right) \right\}_{r \in \mathcal{R}} \right),
\end{align}
where $\mathcal{N}_r^t(v)$ denotes the neighbors of node $v$ under relation $r$ at time $t$. DyHAN~\cite{yang_dynamic_2020} and HTGNN~\cite{fan_heterogeneous_2022} implement this via hierarchical attention, aggregating first across neighbors of the same type, then across types, and finally over time. Some approaches also incorporate hypergraph structures. For example, HTHGN~\cite{liu_heterogeneous_2025} generalizes the receptive field to high-order group interactions through heterogeneous hyperedge construction and expansion to preserve local connectivity while capturing group dynamics.

Despite their expressiveness, such hierarchical architectures are difficult to scale, as separate message-passing over neighbor types and historical snapshots incurs substantial computational and memory overhead. Moreover, repeated attention-based aggregation across both type and time dimensions potentially leads to over-smoothing.

\subsubsection{Meta-structure-guided GNNs}
Meta-structure-guided GNNs predefine meta-paths or meta-graphs based on specific semantics and constrain message-passing only between nodes reachable from  these structures, thus explicitly modeling semantic dependencies. Formally, given a snapshot $\mathcal{G}^{(t)}$ and a meta-structure $\mathcal{S}\in\{\mathcal{P},\mathcal{M}\}$, the representation of node $v$ is computed by aggregating messages from its meta-structure-induced context $\mathcal{N}_s^{(t)}(v)$:
\begin{equation}
\mathbf{h}_v^{(t)} = \text{AGG}_{\text{time}} \left( \left\{ \text{AGG}_{\text{meta}} \left( \{ \mathbf{h}_u^{(t)} \}_{u \in \mathcal{N}_s^{(t)}(v)} \right) \right\}_{s \in \mathcal{S}} \right).
\end{equation}
HDGAN~\cite{li_heterogeneous_2020} constrains propagation by enumerated meta-paths and treats each meta-path as an independent semantic channel, enabling additive composition of path-level semantics via attention-based fusion. DHANE~\cite{li_dynamic_2024} merges multiple paths into a unified meta-graph structure and performs message-passing at the meta-graph level, thereby capturing cross-path contextual dependencies in a single propagation process. Furthermore, MG-DVD~\cite{liu_mg-dvd_2021} pre-enumerates frequent meta-graphs as semantic templates, under which node or graph representations are aggregated and then aggregated by template-level attention.

While semantically interpretable, these methods suffer from combinatorial explosion when enumerating meta-structures over rich type schemas, and their static semantic templates cannot adapt to temporal shifts in which meta-structures become salient at different evolutionary stages.

\subsubsection{Recurrent-based GNNs}
Recurrent-based GNNs capture the temporal evolution of node representations by maintaining historical states that are recursively updated across snapshots or interactions. The core mechanism employs a heterogeneous GNN to encode structural information at each snapshot, coupled with recurrent units to evolve the node memory $\mathbf{M}^{(t)}$:
\begin{equation}
\mathbf{m}_v^{(t)} = f_{\text{RNN}} \left( \mathbf{m}_v^{(t-1)}, f_{\text{HGNN}} \left( G^{(t)}, \mathbf{X}^{(t)} \right) \right), \\
\end{equation}
where $\mathbf{m}_v^{(t)}$ denotes a recurrent memory state of node $v$ at time $t$, and $f_{\text{HGNN}}(\cdot)$ encodes the heterogeneous structure and features of snapshot $\mathcal{G}^{(t)}$.
DyHATR~\cite{hutter_modeling_2021} employs a hierarchical attention mechanism as the encoder to capture heterogeneity within each snapshot, followed by an RNN and temporal self-attention to model sequential dependencies. 
DURENDAL~\cite{dileo_durendal_2023} generalizes this paradigm by investigating the structural placement of the recurrent update relative to the heterogeneous aggregation, proposing both Aggregate-Then-Update and Update-Then-Aggregate schemes to balance memory efficiency and relational temporal dynamics.
Unlike the above methods that separate structural encoding and temporal updating, HGN2T~\cite{liu_hgn2t_2024} proposes a plug-and-play coupling mechanism that tightly integrates a static heterogeneous GNN with a recurrent module, allowing the temporal evolution of node representations to be conditioned jointly on heterogeneous graph structure and relation-specific dynamics.

In summary, GNN-based methods enable end-to-end learning with expressive message-passing, yet remain constrained by hierarchical decomposition that prevents parallelization, snapshot discretization that discards fine-grained temporal signals, and computational complexity that scales multiplicatively with relation types and temporal depth.

\begin{table*}[!htb]
\centering

\resizebox{\textwidth}{!}{%
\begin{tabular}{cccccc}
\hline
Category & Method & Granularity & Core Technique & Task & Code \\ \hline
\multirow{10}{*}{\rotatebox[origin=c]{90}{Embedding-based}}
& Change2vec~\cite{bian_network_2019} & Dis. & Meta-path RW with Incremental Update & Clust. & \href{https://github.com/Change2vec/change2vec}{Link} \\
& THINE~\cite{huang_temporal_2021} & Con. & Chronological Meta-path Walk with Hawkes Process & NC/LP/TLP & \href{https://github.com/S-rz/THINE}{Link} \\
& DyHINE~\cite{xie_learning_2021} & Dis. & Relation-specific RW with Att. & TLP & - \\
& HPGE~\cite{ji_dynamic_2021} & Con. & Relation-specific Att. with TPP & NC/TLP & \href{https://github.com/BUPT-GAMMA/HPGE}{Link} \\
& M-DHIN~\cite{fang_scalable_2022} & Dis. & Meta-graph Sampling with LSTM & NC/LP/TLP & - \\
& DyHNE~\cite{wang_dynamic_2022} & Dis. & Meta-path Proximity with Eigen Perturbation & NC/RP/TLP & \href{https://github.com/rootlu/DyHNE}{Link} \\
& LIME~\cite{peng_lime_2022} & Dis. & Meta-path RW with Recursive NN & NC/Clust. & \href{https://github.com/RingBDStack/LIME}{Link} \\
& SemE~\cite{zhou_temporal_2023} & Con. & Semantic Meta-path Walk with Hawkes Att. & NC/RP/TLP & \href{https://github.com/CGCL-codes/SemE}{Link} \\
& H2TNE~\cite{bai_h2tne_2023} & Con. & Relation-specific Walk with Hyperbolic Embedding & NC/LP/TLP & \href{https://github.com/TaiLvYuanLiang/H2TNE}{Link} \\
& DHANE~\cite{li_dynamic_2024} & Dis. & Meta-graph Att. with Online Update & NC & \href{https://github.com/cs-recommendation/online-recommendation}{Link} \\ \hline

\multirow{8}{*}{\rotatebox[origin=c]{90}{GNN-based}}
& HDGAN~\cite{li_heterogeneous_2020} & Con. & Meta-path GNN with TPP & NC/Clust. & - \\
& DyHAN~\cite{yang_dynamic_2020} & Dis. & Relation-specific GNN with Temporal Att. & TLP & - \\
& DyHATR~\cite{hutter_modeling_2021} & Dis. & Relation-specific GNN with Temporal Att. & TLP & \href{https://github.com/skx300/DyHATR}{Link} \\
& HTGNN~\cite{fan_heterogeneous_2022} & Dis. & Relation-specific GNN with Temporal Modeling & NR/TLP & \href{https://github.com/YesLab-Code/HTGNN}{Link} \\
& DURENDAL~\cite{dileo_durendal_2023} & Dis. & Relation-specific GNN with RNN & LP/TLP & \href{https://anonymous.4open.science/r/durendal-5154}{Link} \\
& HGN2T~\cite{liu_hgn2t_2024} & Dis. & Relation-specific GNN with GRNN & LP/TLP & \href{https://github.com/huanliucs/HGN2T}{Link} \\
& HTHGN~\cite{liu_heterogeneous_2025} & Dis. & Relation-specific GNN with Temporal Att. & LP/TLP & \href{https://github.com/huanliucs/HTHGN}{Link} \\
& TeSa~\cite{wang_tesa_2025} & Con. & Relation-specific GNN with TPP & LP/TLP & - \\ \hline

\multirow{8}{*}{\rotatebox[origin=c]{90}{Transformer-based}}
& HGT~\cite{hu_heterogeneous_2020} & Con. & Relation-specific Transformer with RTE & TLP & \href{https://github.com/acbull/pyHGT}{Link} \\
& HT-Trans~\cite{9892546} & Con. & Relation-specific Transformer Encoder & LP/TLP & - \\
& DHGAS~\cite{zhang_dynamic_2023} & Dis. & Relation- and Time-aware Self-Att. & NC/NR/RP/LP/TLP & \href{https://github.com/wondergo2017/DHGAS}{Link} \\
& THGAT~\cite{zhang_dynamic_2023-1} & Con. & Relation-specific Att. with RTE & NC/Clust./TLP & \href{https://github.com/linz2000/THGAT}{Link} \\
& STHN~\cite{li_simplifying_2023} & Con. & Structural-Temporal Transformer & LP/TLP & \href{https://github.com/celi52/STHN}{Link} \\
& CasMLN~\cite{wang_llm-enhanced_2024} & Dis. & LLM-enhanced Relation-specific Transformer & NC/NR/RP/LP/TLP & \href{https://github.com/PasaLab/CasMLN}{Link} \\
& HTGformer~\cite{wang_htgformer_2025} & Dis. & LLM-enhanced Transformer for DHG & NC/NR/RP/LP/TLP & - \\
& MIGNN~\cite{yue_unified_2025} & Con. & Att.-based GNN with GRU & LP/TLP & - \\ \hline

\multirow{10}{*}{\rotatebox[origin=c]{90}{Application-oriented}}
& HTGT~\cite{fan_heterogeneous_2021} & Dis. & Relation-specific Att. with RTE & Cybersecurity & \href{https://github.com/kdd2021drdroid/KDD2021_DrDroid/tree/main}{Link} \\
& ATHITD~\cite{qi_athitd_2025} & Dis. & Relation-specific Att. with RTE & Cybersecurity & - \\
& HetETA~\cite{hong_heteta_2020} & ST. & Heterogeneous GNN with Causal Conv. & Traffic & \href{https://github.com/didi/heteta}{Link} \\
& STHGFormer~\cite{li_towards_2024} & ST. & Relation-specific Att. with Temporal Att. & Traffic & - \\
& SUPA~\cite{wu_instant_2023} & Con. & Meta-path RW with Memory Update & Recomm. & \href{https://github.com/NoMultiply/SUPA}{Link} \\
& DHGP~\cite{li_periodic_2024} & Dis. & Relation- and Time-aware Att. & Recomm. & \href{https://github.com/AllminerLab}{Link} \\
& DHGRNN~\cite{wang_two-stream_2025} & ST. & Relation-specific Att. with GRU & Multimodal & - \\
& HDGR~\cite{dai_cross-modal_2025} & Dis. & Relation-specific GConv. with Reconstruction & Multimodal & - \\
& SI-HDGNN~\cite{xu_heterogeneous_2022} & Con. & Relation-specific RW with GRU & Diffusion & \href{https://github.com/xovee/si-hdgnn}{Link} \\
& THGNN~\cite{li_thgnn_2023} & Con. & Relation-specific Att. with LSTM & Diffusion & \href{https://github.com/TongjiFinLab/THGNN}{Link} \\ \hline
\end{tabular}
}
\caption{Comparison of DHG representation learning methods.
Task abbreviations: NC (node classification), LP/TLP (link / temporal link prediction),
RP (relation prediction), NR (node regression), Clust. (node clustering).
Temporal granularity: Dis. (discrete-time), Con. (continuous-time), ST. (spatio-temporal).
Technique abbreviations: RW (random walk), Att. (attention),
GConv. (graph convolution), GRNN (graph recurrent neural network),
TPP (temporal point process), RTE (relative time encoding).}
\label{tab:method_comparison}
\vspace{-3mm}
\end{table*}

\subsection{Transformer-based}
Transformer-based approaches represent the most recent paradigm shift in DHG representation learning, which leverage architectural advances that have driven breakthroughs across deep learning. By unifying DHG modeling via self-attention, they bypass the hierarchical decomposition inherent to GNN-based methods while enabling parallelized training. Three paradigms operate at distinct granularities: (1) structure-oriented methods, which attend over spatial neighborhoods at snapshot granularity with type-parameterized weights and relative time encodings; (2) interaction-oriented methods, which model each node's event history at interaction granularity through sequence-level attention; and (3) LLM-enhanced methods, which inject external semantic knowledge from LLMs to complement structural representations.

\subsubsection{Structure-oriented Transformer}
Structure-oriented Transformers unify heterogeneous and temporal modeling within a single attention mechanism, departing from hierarchical GNNs that aggregate separately across relation types and time steps. The core paradigm computes attention over the full dynamic heterogeneous neighborhood, where attention weights jointly depend on source/target node types, edge type, and relative temporal position:
\begin{equation}
\mathbf{h}_v^{(l)} = \sum_{u \in \mathcal{N}(v)} \alpha_{vu}^{(r)}\cdot f_{\text{MSG}}\left( \mathbf{h}_u^{(l-1)}, f_{\text{RTE}}(\Delta t) \right), \\
\end{equation}
where $\alpha_{vu}^{(r)}$ denotes the heterogeneous and temporal attention weight parameterized by the meta-relation triplet $(\phi(u), r, \phi(v))$ and relative temporal encoding $f_{\text{RTE}}(\Delta t)$. HGT~\cite{hu_heterogeneous_2020} pioneers this paradigm via triplet-based parameter decomposition and sinusoidal relative temporal encoding (RTE).
HT-Trans~\cite{9892546} and DHGAS~\cite{zhang_dynamic_2023} extend this with full Transformer encoders and neural architecture search for attention patterns.
STHGFormer~\cite{li_towards_2024} further refines this by integrating spatio-temporal positional encodings directly into the attention scores.

Despite unifying attention for DHG modeling, structure-oriented Transformers face scalability bottlenecks as attention over full neighborhoods across all relation types incurs quadratic complexity, and fine-grained type parameterization risks overfitting on infrequent meta-relations.

\subsubsection{Interaction-oriented Transformer}
Interaction-oriented Transformers diverge from neighbor-centric structural aggregation by modeling the continuous stream of edge events as a temporal sequence, where self-attention operates over a node's chronological interaction history rather than its spatial neighborhood. The technical core is to maintain a historical event sequence $\mathcal{H}_v = \{e_1, \ldots, e_n\}$ for each node, where each event $e_i = (u_i, v, r_i, t_i)$ carries heterogeneous types and continuous timestamps, and to compute the node representations via sequence-level attention:
\begin{equation}
\mathbf{h}_v^{(t)} = \sum_{i=1}^{|\mathcal{H}_v|} \alpha_i^{(t)} \cdot \left( f_{\text{MSG}}(u_i, r_i) + f_{\text{RTE}}(\Delta t) \right),
\end{equation}
where $\alpha_i^{(t)}$ denotes the attention weight at time $t$ over the $i$-th historical event, $f_{\text{MSG}}(\cdot)$ encodes heterogeneous type information, and $f_{\text{RTE}}(\cdot)$ provides relative temporal position encoding. STHN~\cite{li_simplifying_2023} introduces a unified link encoder that integrates type encoding and relative time encoding, and devises a patching technique that segments long event sequences into fixed-length patches to reduce complexity from quadratic to linear. MIGNN~\cite{yue_unified_2025} further employs a dual-window strategy to separately model short-term fine-grained dynamics and long-term distributional patterns through multi-scale temporal aggregation.

This paradigm excels at fine-grained temporal modeling by capturing event-level dynamics, enabling precise tracking of evolving interaction patterns. However, it incurs memory overhead for maintaining per-node event histories and often underutilizes local structural context.

\subsubsection{LLM-enhanced Transformer}
LLM-enhanced Transformers integrate LLMs to inject external semantic knowledge into DHG learning, addressing the limitations of purely structural methods in capturing implicit type-level properties and domain-specific evolution patterns. Since different node types exhibit distinct temporal dynamics---for example, papers grow monotonically while e-commerce interactions show periodicity---LLMs provide complementary semantic priors that are difficult to learn from topology alone. The core paradigm fuses LLM-derived semantic embeddings with structural representations at multiple granularities:
\begin{equation}
\mathbf{h}_v^{(l)} = \sum_{u \in \mathcal{N}(v)} f_{\theta}\left(\mathbf{h}_u^{(t)}, f_{\text{LLM}}\left(p,\tau (\phi(v), \mathcal{G})\right)\right),\\
\end{equation}
where $p$ indicates the instruction prompt for the current graph context, and $\tau(\phi(v), \mathcal{G})$ denotes the textual description of node type and graph-level characteristics. The fusion function $f_{\theta}(\cdot)$ integrates these complementary views to produce semantically-enriched node embeddings. CasMLN~\cite{wang_llm-enhanced_2024} pioneers this paradigm by constructing structured prompts for both node types and the overall graph, then modulating aggregated representations with LLM-derived embeddings through element-wise product to compensate for skewed heterogeneous data distributions. HTGformer~\cite{wang_htgformer_2025} aligns semantic and structural spaces through instruction tuning, enabling the model to interpret dynamic heterogeneous event patterns as textual instructions, shifting the paradigm from structural encoding towards semantic reasoning.

To summarize, Transformer-based methods unify DHG modeling within attention mechanisms, avoiding hierarchical decomposition. However, they remain constrained by quadratic complexity that limits scalability, additive fusion of type and temporal encodings rather than intrinsic coupling, the absence of pre-training objectives that precludes cross-domain transfer, and underexplored tokenization strategies for heterogeneous node/edge types and temporal granularities.

\subsection{Application-oriented}
Application-oriented approaches formulate domain-specific problems as DHG, serving as a unifying substrate where heterogeneous entities and relations evolve over time. Here we summarize representative applications in the following fields.

\subsubsection{Cybersecurity}
Cybersecurity addresses adversarial behavior detection where attackers deliberately fragment malicious actions across entity types and extended time horizons, rendering individual activities benign in isolation yet collectively revealing coordinated attacks. HTGT~\cite{fan_heterogeneous_2021} models app-market-developer relations through a heterogeneous temporal Transformer that iteratively aggregates spatial dependencies with historical sequences, jointly capturing malware propagation and evolution. MG-DVD~\cite{liu_mg-dvd_2021} translates API event streams into dynamic graphs with discriminative meta-graphs, enabling real-time variant detection via dynamic walks without full retraining. ATHITD~\cite{qi_athitd_2025} introduces temporal neighbors within sliding windows for short-term evolution, while Transformer modules learn long-term drift to highlight anomalous time periods. TimeSAGE~\cite{shekhar_entity_2020} incorporates time-decayed edge weights and temporal random walks to resolve fraudulent identities drifting over subscription lifecycles. These methods require accumulated behavioral evidence, enabling long-horizon attack detection but limiting responsiveness to zero-day exploits. While these methods effectively detect known attack patterns, adapting to adversaries' evolving tactics remains an open challenge.

\subsubsection{Traffic Forecasting}
Traffic forecasting predicts dynamics across infrastructure where road segments follow smooth diurnal periodicity, intersections exhibit abrupt phase-dependent transitions, and transit hubs aggregate multimodal demand with distinct volatility patterns that uniform embeddings would collapse. HetETA~\cite{hong_heteta_2020} constructs multi-relational networks with trajectory sequences, employing parallel temporal convolutions for recent, daily, and weekly patterns before graph propagation. STHGFormer~\cite{li_towards_2024} explicitly distinguishes segments from turn nodes via heterogeneous spatial embeddings, with a unified Transformer capturing road-turn interdependencies. REGNN~\cite{luo_dynamic_2020} dynamically constructs event-centric graphs encoding ride requests with supply-demand context, enabling inductive generalization through transferable demand patterns. Although effective in capturing spatio-temporal patterns on observed road networks, generalization to unseen topologies or non-recurrent events remains limited.

\subsubsection{Recommendation}
Recommendation infers preferences from behavioral signals carrying heterogeneous semantic weight and temporal dynamics: clicks indicate transient interest with rapid decay, purchases reflect committed preferences with lasting influence, and reviews encode explicit sentiment requiring delayed integration. DHIMN~\cite{xie_sequential_2021} maintains type-specific memory states with two-level attention that weights items within behavior types then aggregates across types. DHGP~\cite{li_periodic_2024} introduces periodic prompts aligning cyclic interests with temporal context through prompt embeddings modulating basket representations. SUPA~\cite{wu_instant_2023} addresses neighborhood disturbance via sampling influenced subgraphs, updating endpoints, and propagating incrementally without full retraining. Despite effectively capturing type-specific preference, these methods remain limited under cold-start conditions with sparse heterogeneous interactions.

\subsubsection{Multimodal Learning}
Multimodal learning integrates data streams whose semantic alignments shift as context evolves, particularly when modalities exhibit asynchronous sampling rates or intermittent availability that preclude fixed correspondence assumptions. HSSHG~\cite{wang_hsshg_2024} constructs spatio-temporal graphs where co-occurrence and consistency constrain adjacency, using plot summaries and locations as priors to weight edges for video question answering. DHGRNN~\cite{wang_two-stream_2025} fuses spatial-temporal and spatial-spectral streams, with graph Transformers modeling channel heterogeneity and evolving convolutions adapting to missing EEG channels as structure changes. HDGR~\cite{dai_cross-modal_2025} builds dynamic intra-modal graphs with bipartite cross-modal graphs capturing semantic transitivity for 3D retrieval. While effective in fusing multimodal data with dynamic alignments, robustness to missing or noisy modalities remains underexplored.

\subsubsection{Information Diffusion}
Information diffusion examines propagation across heterogeneous actors governed by type-specific dynamics. For example, citations accrue monotonically, evidencing durable scholarly influence; social endorsements decay rapidly in accordance with attention cycles; and financial contagion transmits through correlation structures that shift abruptly under changing market regimes. HINTS~\cite{jiang_hints_2021} addresses cold-start prediction by imputing pseudo-historical trajectories via network alignment, converting embeddings into citation model parameters. SI-HDGNN~\cite{xu_heterogeneous_2022} models academic influence via temporal-attentive aggregation over directed author-paper-venue graphs. NetCycle+~\cite{xiong_netcycle_2018} incorporates life-cycle stages, recognizing that nodes at different evolutionary phases exhibit distinct patterns. THGNN~\cite{li_thgnn_2023} constructs daily correlation graphs with heterogeneous attention for financial contagion. Despite these efforts, unifying diffusion dynamics across actor types with disparate temporal scales remains open.

\section{Open-source Datasets and Benchmarks}
This section summarizes commonly used open-source DHG datasets, covering their application domains, numbers of node and edge types, number of time snapshots/timestamps, availability of node labels, and data source links. The datasets are ordered in descending usage frequency in Table~\ref{tab:datasets}.

\begin{table}[bp]
\centering
\resizebox{\columnwidth}{!}{%
\begin{tabular}{cccrrrcc}
\hline
\multicolumn{1}{l}{} & Domain & Dataset & $|\mathcal{A}|$ & $|\mathcal{R}|$ & $|\mathcal{T}|$ & Label & URL \\ \hline
\multirow{10}{*}{\rotatebox[origin=c]{90}{Datasets}}
 & Academic & DBLP & 4 & 3 & 10 & $\checkmark$ & \href{https://dblp.org/}{Link} \\
 & Academic & AMiner & 4 & 3 & 10 & $\checkmark$ & \href{https://www.aminer.cn/open/article?id=655db2202ab17a072284bc0c}{Link} \\
 & Review & Yelp & 4 & 3 & 10 & $\checkmark$ & \href{https://www.yelp.com/dataset}{Link} \\
 & Social & Twitter & 1 & 3 & 7 & $\times$ & \href{http://snap.stanford.edu/data/higgs-twitter.html}{Link} \\
 & Epidemic & COVID & 2 & 3 & 304 & $\checkmark$ & \href{https://coronavirus.1point3acres.com/}{Link} \\
 & Recomm. & Movielens & 3 & 3 & 25,864 & $\times$ & \href{https://www.kaggle.com/datasets/grouplens/movielens-20m-dataset}{Link} \\
 & Recomm. & Ecomm & 2 & 4 & 11 & $\times$ & \href{https://tianchi.aliyun.com/competition/entrance/231719}{Link} \\
 & Academic & MAG & 4 & 4 & 10 & $\checkmark$ & \href{https://ogb.stanford.edu/docs/nodeprop/}{Link} \\
 & Recomm. & Alibaba & 2 & 4 & 11 & $\times$ & \href{https://tianchi.aliyun.com/competition/entrance/231719}{Link} \\
 & Academic & Math-Overflow & 1 & 3 & 11 & $\times$ & \href{http://snap.stanford.edu/data/sx-mathoverflow.html}{Link} \\ \hline
\multirow{4}{*}{\rotatebox[origin=c]{90}{TGB 2.0}} 
 & Software & software & 4 & 14 & 689,549 & $\times$ & \href{https://tgb.complexdatalab.com/docs/thg/\#thgl-software}{Link} \\
 & Social & forum & 2 & 2 & 2,558,457 & $\times$ & \href{https://tgb.complexdatalab.com/docs/thg/\#thgl-forum}{Link} \\
 & Software & github & 4 & 14 & 2,510,415 & $\times$ & \href{https://tgb.complexdatalab.com/docs/thg/\#thgl-github}{Link} \\
 & Recomm. & myket & 2 & 2 & 14,828,090 & $\times$ & \href{https://tgb.complexdatalab.com/docs/thg/\#thgl-myket}{Link} \\ \hline
\end{tabular}%
}
\caption{Overview of Open-source DHG Datasets and Benchmark.}
\label{tab:datasets}
\end{table}

Note that although many studies nominally rely on the same public sources, they frequently employ inconsistent graph construction rules, temporal discretizations, and evaluation protocols, which undermine reproducibility and cross-method comparability. To mitigate this, TGB~2.0~\cite{gastinger_tgb_2024} introduces standardized DHG datasets with unified preprocessing and evaluation pipelines, providing a more consistent foundation for benchmarking and facilitating future extensions. Nevertheless, the current release primarily targets link prediction and includes only a relative limited set of baseline methods. Prevailing fixed-split evaluation protocols that test solely on final-period edges fail to capture DHG evolution and tend to overestimate.
Extending task coverage, establishing incremental learning protocols, and dveloping comprehensive benchmarks remain critial for future research. 

\section{Conclusion and Future Directions}

This survey presents the first systematic review of DHG representation learning. We provide a generalized definition of DHGs, establish an algorithm-centric taxonomy, and summarize key modeling biases. Our analysis of representative methods and benchmarks reveals important limitations in current modeling mechanisms and evaluation protocols. Despite recent advances toward foundation models~\cite{liu_graph_2025_tpami}, three critical challenges in DHG representation learning persist: (1) Prohibitive computational complexity hinders scaling to Web-scale graphs, precluding systematic investigation of scaling laws; (2) Current task-specific paradigms lack pre-training frameworks for learning transferable representations across heterogeneous schemas and temporal distributions; (3) Limited interpretability impedes high-stakes deployment and obscures whether models capture causal mechanisms versus spurious correlations. We conclude by identifying three progressive research frontiers to address these challenges that will catalyze future advances in this rapidly evolving field.

\paragraph{Efficiency.}
The quadratic complexity of current DHG learning methods~\cite{hu_heterogeneous_2020,li_towards_2024} that explicitly couple message-passing with temporal aggregation fundamentally restricts their applicability to Web-scale DHGs. While recent efforts have attempted to alleviate this bottleneck through partial linearization strategies such as event patching~\cite{li_simplifying_2023} or incremental propagation~\cite{fang_scalable_2022},  the expressiveness-efficiency trade-off persists. Thus, a critical open problem is to explore efficient alternatives, such as kernel-approximation-based variants of linear attention mechanisms or sparsification strategies for DHGs, as a prerequisite for investigating the scaling laws of DHG representation learning.

\paragraph{Generalizability.}
Current DHG methods are largely confined to task-specific supervised training, failing to exhibit transfer capabilities seen in vision and language foundation models. Although preliminary attempts have integrated LLM-derived semantics~\cite{wang_llm-enhanced_2024} or unified attention mechanisms across heterogeneous types~\cite{wang_htgformer_2025}, the field still lacks principled pre-training objectives for DHGs. Future research could develop pre-training paradigms such as self-supervised tasks, combined with transfer learning, few-shot adaptation, and continuous learning, to enable model generalizability across multi-domain evolutionary patterns.

\paragraph{Trustworthiness.}
As DHG models increasingly inform high-stakes decision-making, interpretability is imperative for ensuring trustworthiness. Current  approaches~\cite{li_heterogeneous_2023} emphasize post-hoc subgraph identification but lack faithfulness guarantees and causal grounding. Future work should prioritize inherently interpretable architectures, such as attention or bottleneck layers projecting onto human-interpretable concepts, and integrate structural causal models with DHG learning to distinguish genuine causal relationships from spurious correlations across heterogeneous entities and time.


\section*{Acknowledgments}
Pengfei Jiao was partially supported by the National Natural Science Foundation of China under Grant No.~62372146, the Zhejiang Province Key R\&D Program Project under Grants No.~2024C01212 and 2025C01023, and the Zhejiang Provincial Key Laboratory for Sensitive Data Security Protection and Confidentiality Management under Grant No.~2024E10048. Jie Yin was partially supported by the Australian Research Council under Grant No.~DP250100871.

\bibliographystyle{named}
\bibliography{ref}

\clearpage

\end{document}